***Chapter 7.*** Change Detection for Geodatabase Updating

Rongjun Qin
Department of Civil Environmental and Geodetic Engineering
Department of Electrical and Computer Engineering
Translational Data Analytics Institute
The Ohio State University
2036 Neil Avenue, Columbus, Ohio, USA

[Part of the contents are exceptions from the author's Ph.D. thesis "3D Change Detection in an urban environment with multi-temporal data" and review paper "3D change detection – Approaches and applications"]

Abstract: The geodatabase (vectorized data) nowadays becomes a rather standard digital city infrastructure; however, updating geodatabase efficiently and economically remains a fundamental and practical issue in the geospatial industry. The cost of building a geodatabase is extremely high and labor intensive, and very often the maps we use have several months and even years of latency. One solution is to develop more automated methods for (vectorized) geospatial data generation, which has been proven a difficult task in the past decades. An alternative solution is to first detect the differences between the new data and the existing geospatial data, and then only update the area identified as changes. The second approach is becoming more favored due to its high practicality and flexibility. A highly relevant technique is change detection. This article aims to provide an overview the state-of-the-art change detection methods in the field of Remote Sensing and Geomatics to support the task of updating geodatabases. Data used for change detection are highly disparate, we therefore structure our review intuitively based on the dimension of the data, being 1) change detection with 2D data; 2) change detection with 3D data. Conclusions will be drawn based on the reviewed efforts in the field, and we will share our outlooks of the topic of updating geodatabases.                      .

## 1. Introduction

### 1.1. Background

The world never remains static. The moment a geodatabase (vector data) was constructed (normally at a high cost), a substantially practical problem follows: how these data can be updated economically and timely given the fast development of land processes and dynamics of the man-made objects. In many cases, the maps (geodatabase) we use have at least several months of latency. However, the current approaches, for example in updating the basemap of the car navigation system, relies on hiring thousands of labors driving at every street/road in the urban/suburban/rural areas to manually record, verify and regenerate the maps, which is extremely time consuming and expensive. This intuitively points the solution path to approaches that construct geodatabase more automated, however, this is indeed a long-existing research problem that requires technical breakthrough in automated feature extraction, object extraction and topological reconstruction, which unfortunately have been largely unsolved problems for many decades (in terms of application of engineering-grade data generation). An alternative solution - change detection, proposes not to focus on database construction speed and automation, but to identify potential changed areas efficiently and only to update them. This is apparently related to the development of the well-known change detection problems, which performs a direct comparison of multi-temporal data to yield temporal differences for geodatabase updating. The change detection and updating approach (sometimes called change modeling, which is considered identical to change detection in this article) is becoming a favorable trend in addressing geodatabase updating problem, mainly due to several advantages:

1) In comparison to the approaches of developing more automated methods in geospatial data extraction, change detection approaches can significantly reduce amount of data being remodeled and has the potential to be deployed in a relatively short time.

2) The change detection algorithms normally aim to automatically detect the differences of data, this reflects a much faster identification of the dynamics of ground object, and the change itself can be used as an independent product;

3) there are plenty of timely (and sometimes free) multi-temporal data sources available, with levels of detail ranging from continental scale to city and individual object scale, allowing the development of flexible multi-resolution strategies for change detection to further reduce the cost.

*1.2. Scope of this paper*

Change detection (CD) in remote sensing and Geomatics refers to the process of identifying differences in the state of interested objects or phenomenon by observing it at different times (Singh, 1989). It is the most important step for updating the geodatabase, where the updating itself is followed by a reconstruction (vectorization) of the changed object, which uses common approaches (manual or semi-automated). This paper aims to provide an overview of the start-of-the-art approaches of CD and geodatabase updating. CD is a well-defined and classic topic in the field of remote sensing, with the aims of detecting the differences and monitoring the land process and dynamics in a relative coarse level. This paper will mainly focus on methods that adopts remote sensing and Geomatics data for geodatabase updating at various level of details. Other relevant research topics, such as real-time video sequence analysis, multi-temporal data analysis and indoor/outdoor object tracking, partially regarded as a process of change detection, will be out of the scope. In this review, we primarily consider change detection approaches in bi-temporal data (before and after), and methods using of multi-temporal data will be occasionally presented while not inclusive.

The types and formats of data may vary greatly, and there are generally two types of data: raster data and vector data. To be specific, raster data are unstructured raw data that come directly from the sensors or are only through preliminary low-level processing (e.g. de-noising, format conversion, decoding), and examples are images, stereo/multi-stereo image blocks, filtered/raw 3D point clouds, and 3D triangular meshes. Vector data refers to structured data where necessary information have been extracted, such as DLG (Digital Line Graphs), 3D polyhedral models, building footprints with associated information, e.g. number of floors of the building, area/type of the buildings. Updating vector data using images/3D point clouds is directly relevant, while methods that purely use bi-temporal images/point clouds are equally relevant, as the extracted changes can be directly applied to the geodatabase for updating. In our paper, we consider the review of change detection methods in the field of remote sensing and Geomatics, while limiting to methods that use 1) Unstructured raw data: optical images, LiDAR (Light Detection and Ranging) point clouds, triangular photo meshes from airborne, satellite and terrestrial platforms; 2) structured data: DLG, 3D polyhedral models (LoD models). Other data types, such as sonar, microwave radar, and ground penetrating radars, are out of the scope.

*1.3.Organization of this paper*

CD is a highly disparate problem that varies greatly with the data quality, resolution and format. This leads to many different types of categorizations of CD methods, e.g. based on the basic processing unit, data types and processing methods (Lu et al., 2004; Tewkesbury et al., 2015). This paper adopts a very intuitive classification of methods based on the dimension of the data, being both 1) change detection with 2D data and 2) change detection with 3D data. Following a section (section 2) that briefly introduces the necessary data preprocessing methods, these two types of methods are structured as two main sections (section 3 and 4), in which we will structure the sub-sections based on the predominant methods in the existing literature. In a last section, we share our concluding remarks and outlooks of the topic of geodatabase updating. It should be noted 2D and 3D CD are not completely isolated, for methods that use a mixture of 2D and 3D data (e.g. using DSM to update DLG), we categorize them as change detection with 3D data.

## 2. Data preprocessing

Raw data acquired from different platforms need to be processed into geo-referenced/formatted dataset. Geo-referencing of data acquired from different platforms varies due to data formation and accuracy of the onboard positioning sensors (i.e., GPS (Global positioning System) /IMU (Inertial Measurement Unit)). Geo-referencing of images and LiDAR nowadays becomes routine procedures, with many processing software packages available (HEXAGON, 2014; Pix4D, 2017; Terrasolid, 2013). Satellite data are often coarsely geo-referenced based on the positioning sensors in the spaceborne platform prior to their release. The unstructured data contains images and point clouds, generically representing spectral (color) and geometric information (2D/3D shape). Therefore, before CD, data alignment is needed on both spectrum and geometry, being radiometric correction and geometric co-registration. For CD with stereo/multi-stereo images, DSM generation stage is normally necessary to process the 3D information.

***Radiometric Correction***: Radiometric correction has been a basic process in many image-based CD methods. It can be generally categorized into absolute and relative methods (Hu et al., 2011). Absolute method aims to recover the absolute surface reflectance (Gordon, 1997) based on the radiometric values from the images, while relative correction aims at converting the spectrum (color) of an image to a reference image (sometimes of the same scene) (Roy et al., 2008).

The absolute methods are normally used to deal with low-to-medium resolution (LTMR) remote sensing data, usually through a process of atmospheric correction using radiative transferring models (Berk et al., 1999). The recovered surface reflectance is seen as invariant of sensors and acquisition conditions, such that they can be used for direct comparisons. However, to accurately recover the surface reflectance, many in-situ data are necessary, such as weather, aerosol optical depth, humidity and so on, which are often hard to obtain. Therefore, the absolute methods are only considered when identifying the actual values of surface reflectance is needed, or when they have already been processed through other applications. However, for the purpose of updating geodatabase, the actual surface reflectance values are not necessary.

Practically, the relative methods (relative correction or relative normalization) are more preferred as they do not require additional observations except for the images themselves. This normally refers to the correction of a multiplicative and additive intensity change, and the correction parameters can be estimated either through a few reference pixels or all pixels and patches (El Hajj et al., 2008; Yang and Lo, 2000). The maximum-and-minimum range normalization does not consider the additive intensity change (Yang and Lo, 2000). In the process of estimating the correction parameters (scale and offset, relevant to multiplicative and additive intensity change), blunder pixels such as those saturated by clouds or significantly changed areas should be eliminated (Paolini et al., 2006). One of the recent methods uses spatio-temporal filtering to perform pixel-wise correction (Qin et al., 2016b). This approach applies non-parametric correction to multi-temporal datasets using a 3D bilateral filter. This approach can further be applied to bi-temporal images; however it works more effectively with multi-temporal images.

***Data Co-registration***: Image co-registration refers to the process of aligning the dataset into the same coordinate frame, such that the data from corresponding geographical locations can be compared directly (Chen et al., 2014; Fung, 1990; Fung and Ledrew, 1987). The key for data co-registration is to find sparse or dense corresponding points/areas (e.g., the whole dataset), for applying geometric transformations between two datasets. The correspondences can be either selected manually or extracted using automatic algorithms (Rogan et al., 2002). The co-registration of 2D data assumes that both 2D datasets represent planar spaces, where affine transformation or projective transformation (also called homography) (Hartley and Zisserman, 2003) can be applied. Such approaches work with orthorectified 2D data (parallel projection), however when applied to perspective images, it requires images taken from the same perspective, or nearly nadir for remote sensing images (Bouziani et al., 2010; Pacifici et al., 2009). Moreover, the height relief of the ground objects are non-ignorable for high resolution images and affine assumptions are no longer suitable (Qin, 2014a). A comprising solution would be correcting the distortions using TIN (Triangulated Irregular Network) which is based on sparse (corresponding) points between images (Qin et al., 2013; Wu et al., 2012). Very often remote sensing images are coarsely geo-referenced, therefore they approximately align to a geo-coordinate system. When these images are applied to update the DLG, there are two intuitive approaches:

1) Find correspondences by applying different corner extraction method on both the DLG (vector data) and image data points (either manually or automatically selected) and then apply rigid, similarity or affine transformations (Zang and Zhou, 2007);

2) Perform an object detection algorithm to extract object footprints (e.g. buildings), and then apply binary image matching and/or vector data matching (Bouziani et al., 2010). This often requires the images to be ortho-rectified to a digital elevation model or at-least mean sea level to correct at least part of the project distortions.

Performing co-registration between 3D datasets is advantageous, since normally the 3D data alignment can be well modeled by 3D rigid or similarity transformations. However, the types of 3D data may bring complicated scenarios: there are normally two types of 3D data,

    a) data carrying explicit 3D information such as DSM, 3D point clouds, 3D models, termed 3D-EXP data, and

    b) data carrying implicit information (termed 3D-IMP such as stereo/ multi-stereo images that have potential to generate 3D information (Qin et al., 2016a). Depending on the input bi-temporal data pair (3D-EXP, 3D-IMP or their mixture), the co-registration can be applied either using the imaging sensory geometry (Fischler and Bolles, 1981) or direct 3D transformations. A common approach of co-registering two sets of 3D-IMP data or mixture of 3D-IMP and 3D-EXP is to use a set of GCPs (ground control points) or corresponding 3D/2D feature points, through the process of relative orientation or bundle adjustment (Fraser and

Hanley, 2003; Triggs et al., 2000). In particular, if a large amount of 2D image correspondences are used under a rigorous sensory model, high-accuracy data alignment can be achieved (Qin, 2014b) for bi-temporal and multi-temporal datasets (Qin et al., 2016b). It is normally recommended to co-register two 3D-IMP datasets before converting them to 3D-EXP datasets, since the process of generating 3D-EXP data from 3D-IMP data (e.g. DSM generation from image blocks) may produce errors (Qin, 2014a; Qin, 2014b; Qin and Gruen, 2014).

Methods for co-registering two sets of 3D-EXP data (e.g. 3D point clouds, DSM, triangular meshes) are normally performed through 3D rigid or similarity transformations, using either a set of sparse corresponding points or the whole dataset (global method). The global methods minimize the summed squared error of all the points of two datasets, such as least squares 3D matching (Gruen and Akca, 2005) and Iterative closest point (ICP) algorithm (Besl and Mckay, 1992). These global methods have outlier removal procedures that are robust to data with a certain level of noise. The co-registration of DSM is usually simplified by estimating a 3D shift between two datasets, while terrestrial 3D-EXP data are often more complicated for co-registration due to the complex geometry and occlusions, which might require good approximation values when both datasets are in different coordinate frame.

***DSM Generation***: A necessary step to convert oriented stereo/multi-stereo images into DSMs or point clouds is to perform the dense image matching (DIM). The DIM methods with multiple images can be generally categorized based on how images are structured (Remondino et al., 2014), 1) multi-stereo matching with depth fusion (MSM); 2) Multi-view matching (MVM). MSM is a direct extension of two-view stereo matching, in which images are paired and point clouds of each pair are fused/filtered (in the depth direction) to form final point clouds (Haala and Rothermel, 2012a; Hirschmüller, 2005). MVM considers matching points across multiple images simultaneously (Baltsavias, 1991; Furukawa and Ponce, 2010). MVM is a more rigorous way to incorporate redundant information, but often more complicated to implement. Both types of methods have advantages and disadvantages, and their performances vary with the camera network, scene content, and complexity, strategies for point matching (global or local) (Tao et al., 2001; Yang, 2012). Our own experience is that generally for top-view photogrammetric images blocks (60-80% overlap for frame images and 15-25 degrees of intersection angle for satellite images), the MSM methods such as those using SGM (semi-global matching) appear to be a good choice, it leverages both speed and performances (d'Angelo and Reinartz, 2011; Krauß et al., 2013). However, for terrestrial images, especially for those that form large baselines and poor camera networks, MVM methods in general produce more complete point clouds, since the visibility are modeled while many stereo algorithms tend to resist objects with large parallaxes (Morgan et al., 2010; Seitz et al., 2006).

### 3. Change detection with 2D data

In the remote sensing domain, image-based 2D change detection (CD) has been intensively investigated in the past with the review papers (Coppin et al., 2004; Hecheltjen et al., 2014; Hussain et al., 2013; Jianya et al., 2008; Lu et al., 2004; Radke et al., 2005; Singh, 1986; Singh, 1989; Tewkesbury et al., 2015) and even reviews of review papers covering each stage of the development of CD methods (İlsever and Unsalan, 2012). Image-based 2D CD algorithms are essentially based on image analysis techniques. They aim at detecting the appearing, disappearing and change of the ground objects by analyzing and interpreting the bi-temporal remote sensing images. In most cases, such techniques are applied to images for small-scale change analysis such as urban sprawl and deforestation using LTMR images, with only a small number of them using VHR images. In general, the 2D image-based CD approaches can be categorized into three groups: 1) pixel-based methods, 2) object-based methods and 3) classification-based methods. The first two categories are distinguished by the basic unit for change representation. The classification-based approaches could be either pixel-based or object-based, but the units are interpreted by performing a pre-classification or post-classification. Methods that use images to directly update the vector databases can fall into any of the above three categories, and relevant work will be introduced in each of the three categories.

### 3.1. Pixel-based approaches

Pixel-based CD refers to methods that use pixels as the basic processing unit to check the consistency between two datasets for CD. The class of methods is mostly applied to LTMR images (e.g. Landsat, MODIS, Sentinel), since one pixel is significant enough to represent the whole or relatively large part of the ground, and the mixed-pixel effect (Bioucas-Dias et al., 2012) in this case makes them more robust (less spectrum variation) for a CD in a landscape level (Lambin and Ehrlich, 1997; Lu et al., 2002; Mas, 1999; Metternicht, 1999; Ram and Kolarkar, 1993).

Most of the existing CD methods on LTMR images analyze the discrepancy between image spectral information as well as their transformations. Early works include **visual checking with image overlay, image differencing** (Ingram et al., 1981; Jensen and Toll, 1982), **image ratio** (Howarth and Wickware, 1981; Nelson, 1983), **change vector analysis** (Johnson and Kasischke, 1998; Malila, 1980), **background subtraction**, and **transformation-based method** such as principal component analysis (PCA) (Fung and Ledrew, 1987; Singh, 1989), Tasselled Cap (TC) (Haverkamp and Poulsen, 2003), etc. These methods assumed that changes in the land-cover/land use must result in differences in their radiance values, and these differences should be larger than those caused by other factors such as differences of atmospheric condition, surface moistures, and sun angles. This assumption works for the LTMR images in most of the cases, as these variations are saturated by the mixing pixel effects. In the following, we introduce the predominant methods in pixel-based CD methods.

**Image differencing** refers to the idea of simply subtracting two geo-referenced images with their radiometric values (Muchoney and Haack, 1994; Singh, 1986). It requires preprocessing steps such as absolute/relative radiometric correction and pixel-wise co-registration of the images. After the image differencing, a binary difference map will be generated by applying a pre-defined threshold to the differences, determined through either trial-and-error approaches or statistical measures based on the standard deviation and mean of the image differences. The **Image ratio** computes the fraction between the radiometric values of two images. Researchers have found that using different bands for image rationing could result in different performance (Chavez and MacKinnon, 1994; Howarth and Wickware, 1981; Jensen and Toll, 1982; Nelson, 1983). An advantage of the image rationing method is that it is robustness against radiometric differences due to illumination changes of the environment, which creates higher spectrum separability for CD. The **Change vector analysis** (CVA) on multi-temporal CD stacks the multi-spectral (Johnson and Kasischke, 1998) or multi-level feature differences (Tian et al., 2013) in a vector form. This high dimensional vector is analyzed in its directions and magnitude in the Euclidean space. The direction change can be used to easily infer the significances of bands that contribute to the change (Ye et al., 2016). Each component of the vector can be also weighted according to their significance to the change description for improving detection accuracy. The **Background subtraction** (BS) method assumes that the non-change areas usually have lower spectral/radiometric variances. Prior to the image difference, a low-pass filter is first applied to images to reduce the spatial variances, and then the filtered image is subtracted from the original images (Singh, 1989). This method is easy to implement but has low accuracy, since important spatial features are also smoothed and subtracted.

Direct comparison on spectral bands may result in false alarms due to the ambiguity of the spectral information. Therefore, researchers are interested in **transformation-based** methods for image differencing (Collins and Woodcock, 1994; Fung and Ledrew, 1987; Parra et al., 1996). The main advantage is that the transformed method can reduce the high dimensional features space, and de-correlates inter-band relations (Lu et al., 2004). PCA and TC transformation are orthogonal transformations that transform each component of the original signal to dimensions ordered via the variations. PCA is usually applied to the original image bands and TC is applied to the scene independent components: such as brightness, greenness and wetness (Collins and Woodcock, 1994; Lu et al., 2004; Munyati, 2004). Chen et al., (2013) proposed to use the gradient of the spectrum curve for computing the difference maps, which showed more accurate results than image differencing and CVA method.

A major challenge for 2D CD in LTMR images is the estimation of the signal-to-noise ratio of the images in order to eliminate the unwanted factors induced by the quality of the data. Moreover, even though the LTMR images are supposed to have less perspective distortions, it has been reported that accurate pixel-wise co-registration of images are critical for obtaining reasonable detection accuracies (Lu et al., 2004). Systematic reviews of the 2D CD approaches can be found in (Collins and Woodcock, 1996; Hayes and Sader, 2001; Lu et al., 2004; Radke et al., 2005; Singh, 1989). The sum of squared spectral difference usually results in low separability of real change and unwanted changes, which sometimes are due to the correlation between the images bands. Transformation–based methods aim to remove the inter-bands correlation for a better delineation of changes (Coppin et al., 2001; Coppin and Bauer, 1996; Fung and Ledrew, 1987; Huang et al., 2013; Lillesand et al., 2004; Munyati, 2004). Nowadays the 2D CD techniques on LTMR images become pretty standard, and practical systems are available already in remote sensing software packages such as ERDAS (HEXAGON, 2014) , ENVI (EXELIS, 2014), and eCognition (Trimble, 2014).

### 3.2. Object-based approaches

The objects of interest in VHR images (e.g. IKONOS (1 m), GeoEye (0.5 m), Worldview1/2 (0.5 m), Worldview 3/4 (0.31 m), Pleiades (0.7 m)) may be composed of groups of pixels (Blaschke, 2010), while single pixels with different spectrums may belong to the same object. This leads to the well-known "salt-and-pepper" noises when using pixel-based methods for CD or classification on these images (Yu et al., 2006). Object-based methods are more robust towards this problem: it was noticed in (Fisher, 1997) that a single pixel might not be enough to support the image analysis comparing to pixel groups, and this idea is further supported by (Bontemps et al., 2008; Hay and Castilla, 2008; Hay and Castilla, 2006; Hussain et al., 2013; Longley, 2002). The aim of the object-based methods is to analyze images on a per-object basis, that is, groups of pixels derived following certain criteria (such as textural, spectral/radiometric or semantic homogeneity). These objects can be labeled segments from vector data, or the image segments computed using image segmentation methods (Comaniciu and Meer, 2002). The object-based methods are normally more robust and require less computations (Chen et al., 2012a; Qin and Fang, 2014). In addition, the shape of the segments can be used to identify objects that are not easily differentiable using spectral information along, e.g. building roofs and roads, open ground/plaza and narrow pedestrian paths.

The object-based methods nowadays have been incorporated into major remote sensing image processing packages (EXELIS, 2014; Trimble, 2014) for VHR data. There are several ways of defining an object (Tewkesbury et al., 2015) for bi-temporal dataset:

1) Image-object overlay: the segmentation is performed on one of the images, and the change analysis is performed within each segment (Comber et al., 2004; Listner and Niemeyer, 2011).
2) Image-object comparison: the segmentations are performed on each of the images, change analysis are performed separately and then combined (Boldt et al., 2012; Ehlers et al., 2014).
3) Multi-temporal segmentation: the segmentation is performed on the entire time-series (Bontemps et al., 2012; Teo and Shih, 2013). This includes the case of taking the intersections of individually segmented images (Tian et al., 2013).

Increasingly, studies showed that object-based methods demonstrate improvements over pixel-based methods in classification and CD. Yu et al. (2006) performed a detailed study by classifying vegetation types in airborne images using object-based method, and demonstrated good results towards "salt and pepper" problems. Desclée et al. (2006) proposed an object-based method for change detection on the forest inventory. The intention of using the object-based method was to delineate the change more robustly using statistical testing from pixel values within each object, which reported over 90% of detection accuracy. Conchedda et al. (2008) adopted multi-resolution segmentation techniques on SPOT multispectral image to derive the objects for studying the change of mangrove species. Class-specific rules were derived to incorporate the spectral information in each level of the segmentation, where the mean and standard deviation of the spectral bands of transit classes were used for detecting changes, where high detection accuracy was reported even for small scatters of changes.

CD between 2D images and vector data (from geodatabase) naturally fits the scenario of geodatabase updating. Bouziani et al. (2010) proposed an object-based CD method to automatically detect the building changes between optical images and existing digital cartography data. They compared the extracted segments from the VHR images to the cartographic map based on a set of rules as expert knowledge for Geodatabase updating, and reported 90% of the changes were detected. Similarly, Durieux et al. (2008) proposed an object-based building detection method from SPOT 5 data, and compared it with the existing GIS database to study the urban sprawl phenomenon. Similar methods can be found in (Niemeyer et al., 2008).

Although the advantages of object-based methods are widely recognized, a potential issue comes along is the over- or under- segmentation problem. E.g., for methods that compare the statistics within the objects, under-segmentation may saturate significant changes by incorporating too many unchanged pixels, while over-segmentation may result in changes that are purely induced by noises.

### 3.3. Classification-based approaches

The classification-based approaches generally refer to supervised or unsupervised methods (index-based) that tend to label the pixels/objects before or after the image-based CD. According to its presence order in the CD procedure, it can be further divided into pre-classification (PREC) and post-classification (POSTC) approaches: the PREC methods perform image classification independently for each date, and then take the labeled image as an input for type change identification (Mas, 1999). The POSTC methods first perform image comparison in pixel/object level to find initial change masks, and then discriminate the real changes and unwanted changes (e.g. seasonal varying vegetation) or false positives using classification methods (Pacifici et al., 2007).

The basic argument of the PREC-based approaches is that the direct image comparison (difference of intensities and transformed features, etc.) is sometimes affected by the differences of physical conditions during the data capture. Understanding the scene with learning-based approaches may lead to improved CD accuracy (Al-Khudhairy et al., 2005). Walter (2004) proposed a classification-based method to detect changes between the images and the GIS database. He first employed a maximum likelihood classifier (Foody et al., 1992) to classify the multispectral image in object-level, where the training samples were derived from the GIS database, and then the labeled objects were used for updating the geo-database. A similar method can be found in (Knudsen and Olsen, 2003). Frauman and Wolff (2005) adopted a pixel-based classification method that performed independent classification for multispectral images of each date, and then derived the change maps by checking the consistency between their class labels. Conchedda et al. (2008) proposed to adopt classification-based method to assess the mangrove changes using SPOT XS data, and high detection rate were reported. Similarly, Dronova et al. (2011) applied an object-based nearest neighborhood method to examine the ecosystem change of Poyang lake in China, which showed its values on identifying type transition on ecosystem studies.

The POSTC based CD methods aim to separate real changes from false alarms or unwanted changes using learning-based strategy. Chaabouni-Chouayakh and Reinartz (2011) applied the support vector machine (SVM) learning to separate the changes of trees and buildings from the initial change mask. Pacifici et al. (2007) proposed a two-step classification method for CD, which combined both PREC and POSTC strategy. They first performed a supervised classification using neural-network on the images from different dates independently, and then generated the change maps by comparing their class labels. The two results were then fused by intersecting both change maps.

The classification-based methods can by-pass the uncertainties induced by spectral comparison of the images to a certain extent, since the semantics of each pixel/object are learned from the samples in the current images. However, the CD results are closely related to the classification results. They might vary with the choice of samples, and the granularity of the image segments, which can be sensitive to the parameters and image quality.

## 4. Change detection with 3D data

3D data are becoming increasingly available with reduced cost (Gehrke et al., 2010). Low-cost LiDAR data, more automated acquisition and processing pipeline for airborne, UAV-borne (Unmanned Aerial Vehicles) and space-borne data (Choudhary et al., 2010; Hirschmüller, 2008; Nex and Remondino, 2014; Qin, 2016) are raising interest in change detection (CD) applications in 3D infrastructure monitoring  (Grigillo et al., 2011; Nebiker et al., 2014), volumetric landslide monitoring (Martha et al., 2010), disaster management (Menderes et al., 2015), glacial monitoring (Haala and Rothermel, 2012b), and construction monitoring (Siebert and Teizer, 2014). The major advantages of using 3D data over 2D for CD are threefold:

1) *Insensitive to illumination differences*: The 3D data presents the spatial measurements of the objects, therefore comparison of geometry of bi-temporal data is irrespective of illumination conditions.

 2) *Insensitive to perspective distortions in 2D CD*:  The comparison of geometry can be performed in a true three-dimensional space, or any projected space (subspace of 3D space). Oriented images can be easily corrected based on the available 3D data.

 3) *Volumetric information*: 3D CD provides volumetric changes that can facilitate more applications, such as the volumetric forest loss, accurate construction progress monitoring, etc.

The advantage of 3D data for CD has been known for a long time (Murakami et al., 1999), while 3D CD works are only raised recently, largely driven by the development of automated image-based 3D reconstruction (Remondino et al., 2014). The comparison of bi-temporal data in the height or depth direction obviously leads to more robust results. It is easy to extend some of the 2D CD methods into 3D CD. For instance, pixel/object/classification based methods can all apply to 3D data in a project space (raster forms), such as raster DSM, depth maps. However, one more dimensional information does not necessarily define 3D CD as a simple extension of 2D CD, given more complicated issue it brings in:

1) *Uncertainties of 3D data*: 3D data generation (e.g. from images) brings different types of uncertainties, associated with the image quality, convergence angle, DIM algorithms and object scale. For example, the image matching may fail on thin and tall objects or large texture-less area. Uncertainties of point clouds generated using different dense matching methods may have different and non-uniform distributions.

2) *Multi-modal data fusion:* Geometric data presents a different modality from the image data. Fusion of both data requires special considerations of different types of data uncertainties, feature extraction and multi-source weighting (Tian et al., 2013).

3) *CD under a true 3D environment:* 2D CD methods can be straightforwardly extended to process 2.5D data such as bi-temporal raster DSMs or depth maps. However, in a truly 3D environment (including height and façade reliefs, sometimes non-convex and non-watertight), e.g. 3D data generated from oblique imagery, LiDAR point clouds from mobile/terrestrial laser scanning, 2D CD methods can hardly solve the problem. Moreover, presence of occlusions, disturbances of unwanted objects, incomplete data, and 3D feature extraction in such scenarios, require new CD techniques and methods.

Technically speaking, although this additional dimension brings more complex scenario for CD, there are two fundamental aspects that can simply encapsulate the particularities of 3D CD techniques, being 1) geometric comparison; 2) geometric-spectral analysis. Geometric comparison refers to methods that measure geometric differences of 3D dataset for change determination, while geometric-spectral analysis considers both geometry and spectrum (color) fusion for CD. These two aspects can also be incorporated into a single 3D CD framework. In addition, the 2D image-based (raster-based) techniques can be valuable reference for processing 3D data in a projected space (2.5D data, e.g. height, depth maps). In this section, we largely follow our previous review in (Qin et al., 2016a) and focus on techniques that perform geometric differencing under difference scenarios, as well as methods for fusing both geometric and spectral data for CD.

### 4.1. Geometric Comparison

The geometric comparison varies greatly with the viewing scenario (oblique-view, top-view) and data format (DSM, point clouds, stereo images, etc.). It refers to a 2.5D comparison such as height/depth difference (shown in Figure 1a), or a full 3D comparison through a Euclidean distance measure (shown in Figure 2b). Moreover, image sets taken from different perspectives implicitly contain 3D geometric information (3D-IMP), and the geometric difference of such data requires image comparison through projection (projection-based method) (an example is shown in Figure 1c), or multi-ray consistency evaluation. There is not a best out of the three, each can be applied to an appropriate context depending on the viewing scenario and data formats.

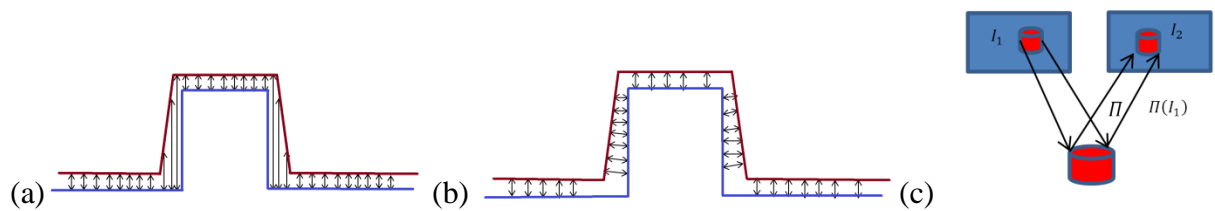

*Figure 1*. Different geometric comparison methods. (a) Height difference, distances are computed vertically. (b) Euclidean distances, distances are computed in the surface normal direction. (c) projection-based inter-correlation method, the geometric difference is computed by projecting image $I_1$ on to the object, and then back project to image $I_2$ as $\Pi(I_1)$; the differences are given by measuring the differences between $\Pi(I_1)$ and $I_2$.

**_Height differencing:_** height differencing is the most basic and simplest type of geometric comparison, usually applied to bi-temporal raster-DSM or depth maps. This essentially treats the height differencing as a raster comparison problem, where the pixel/object based differencing methods can be directly used. Per-pixel height differencing has been widely used in applications such as tree growth monitoring (Gong et al., 2000; Stepper et al., 2015; Waser et al., 2008; Waser et al., 2007), earthquakes and damage assessment (Strong, 1974; Turker and Cetinkaya, 2005) etc. It was also applied to urban areas. Sasagawa et al. (2013) applied the height differencing in the urban area using DSMs generated from ALOS (Advanced Land Observation Satellite) triplets to indicate changes on individual buildings. Due to the presence of noises in the DSM or depth maps, window-based or object-based methods can be used to average the height differences (Tian et al., 2010; Tian et al., 2013). These objects can be derived from available orthophotos associated with the DSM (Tian et al., 2013), as the images normally contains richer information of object boundary. Moreover, refinements can be performed using additional features such as geometric primitives, textural/spectral features, or external data sources such as from GIS (Geographical Information System) database (Dini et al., 2012).

The height threshold, similar to thresholds of image differencing, is one of the most important parameters to determine the final changes. The selection of the threshold is correlated with the accuracy of the DSM. One way for threshold determination is to use a priori information such as the pre-assessment of the DSM quality and empirical choices, or trial-and-error tests (Lu et al., 2004; Murakami et al., 1999). Another way is to estimate the threshold based on statistical measures, such as from the histogram of the height residuals (Chaabouni-Chouayakh and Reinartz, 2011). The height threshold reflects the actual metric of objects, it is however very often data-dependent, which is again due to the different level uncertainty of the bi-temporal 3D data. To avoid single threshold truncation, multiple thresholds can be also used to indicate different levels of confidence (Qin et al., 2015). Regions with a very high confidence of being changed can be used directly as the CD output, while uncertain ones could be sent for operator's decision.

***3D Euclidean distances:*** The Euclidean distance between two surfaces takes the three degrees of freedom (of the 3D geometry) into account by computing the distance along the normal direction, which is theoretically more rigorous. The difference between the Euclidean and height distances can be easily understood in Figure 1(a-b). The Euclidean distance accounts better for the misregistration/DSM errors in the building boundaries by considering the normal direction. Techniques in this category are generally developed in the domain of surface co-registration and CD, where in surface co-registration, changes are regarded as outliers. An example of such technique is the least squares 3D surface matching proposed by (Gruen and Akca, 2005), which performs 3D data co-registration using Euclidean distance measures. It was later applied by (Waser et al., 2008) to estimate the forest volume dynamics between two image-derived DSMs. Under the context of 3D model quality control, Akca et al., (2010) adopted the LS3D method to detect the 3D geometric modeling error against the LiDAR measurements, where Euclidean differences were estimated by minimizing the Euclidean distances between the 3D data. For similar methods for CD, the readers may refer to the global methods that minimize point-to-surface or surface-to-surface distances, where the outliers of the co-registration were detected as changes (Habib et al., 2005; Karras and Petsa, 1993; Maas, 2000; Mitchell and Chadwick, 1999; Pilgrim, 1996; Rosenholm and Torlegard, 1988).

Occlusions and incompleteness of 3D data (from LiDAR or images) present much more complex scenarios than remote sensing top-view data. Akca (2007) showed various successful CD examples using Euclidean distance measure in close range applications under the context of deformation analysis and quality control (Akca et al., 2010). Other derivative measures based on Euclidean distance can be also used for CD: (Girardeau-Montaut et al., 2005) applied an octree structure to divide the 3D spaces, and the Hausdorff measure was employed to compute the distance between different spaces. Similarly, Kang and Lu (2011) adopted the Hausdorff distance to detect the difference between LiDAR scanning data and a reference 3D model.

Sometimes the Euclidean distance between extracted geometric features appears to be more robust than per-point Euclidean distances: Eden and Cooper (2008) measured the differences of 3D lines across two multi-view image sets, which significantly reduced the noise and disturbances. Under the same concept, Champion et al. (2010) extracted 3D lines from stereo images to verify the existence of buildings by comparing them to the GIS database. Nevertheless, failing to detect such features may omit some important changes. Therefore, feature-based methods should only be applied under the context that the object of interest can be represented by certain features.

***Projection-based geometric differences:*** It is not always the best recommendation to generate 3D point clouds/DSM from stereo/multi-stereo images before the CD step. Stereo images taken under suboptimal conditions (poor intersection angle and high environmental illumination differences) may produce unreliable 3D information. If relatively reliable DSM or point clouds are available at one date and images are oriented with respect to the 3D data, the projection-based geometric difference can be used to assess the geometric consistency without explicitly generating 3D data from the images. It correlates, one image of the stereo pair with the other image using the DSM or point cloud, and compares their radiometric/spectral differences (shown in Figure 1c). In principle, these two correlated images should be the same if the stereo pair is consistent with the DSM/point clouds. Qin (2014a) applied inter-correlation in the process of 3D model updating , where two stereo satellite images were correlated using 3D polygonal models, and the correlated image patches were evaluated using the energy produced by SGM (Semi-global matching) algorithm (Hirschmüller, 2008). In (Knudsen and Olsen, 2003), 3D models were projected onto 2D photos, followed by supervised classification for CD.

This technique is particularly effective for the oblique-view images and point clouds/3D models, as a direct comparison using point clouds generated from such images via DIM usually produces many artifacts. Taneja et al. (2011) applied inter-correlation of a stereo pair to an image-derived surface model, and the differences in color were used as change evidence. Qin and Gruen (2014) extended inter-correlation to a multi-stereo case in order to determine view-based change evidences by comparing a strip of images with mobile LiDAR point clouds.

Another streamline of the projection-based method divides the 3D spaces into voxel/object representations, where photo-consistencies of multi-view images projected to the cells are evaluated statistically. Voxels with significant color variations will be spotted as changes, examples are (Crispell et al., 2012; LeCun et al., 2015; Pollard and Mundy, 2007; Schindler and Dellaert, 2010). Such color-consistency check implicitly applies a multi-ray point matching strategy, where false positives might be present in occluded areas and false negatives might occur in non-texture areas. After the probability assignment, the Markov interfering processes (Blake et al., 2011) were applied very often to reduce noise effects.

The projection-based method is an effective strategy to provide raw change evidence when the 3D scene is rather complex, e.g. oblique data or street-view data. It can be seen as an inverse operation of matching, while this again, still depends on the quality of the available 3D data and may not be able to handle areas with insignificant texture features.

### 4.2. *Geometric-spectral Analysis*

Normally the geometric 3D data comes along with spectral/intensity information, e.g. orthophoto associated with the DSM, textures associated with the 3D models. It is straightforward to understand that additional channels of information may lead to enhanced CD results, as the geometric and spectral information could be beneficial to each other, while on the other hand, it faces the risk of error propagating from both sources. Therefore, the main challenge in this case lies in the information fusion for optimal signal-noise ratio (utilize useful information of both sources while suppress their errors). In general, there are three ways to integrate the geometric and spectral features into a 3D CD process:

1) Post-refinement.

2) Direct feature fusion.

3) Classification-based method.

Post-refinement refers to methods that perform refinement on initial change evidences from geometry and/or spectrum comparison (Chaabouni-Chouayakh et al., 2011). The second approach jointly fuses geometric and spectral information (or their transformed features) in the feature level, and the fused features are used to determine the presence of change. The third approach is very similar to classification-based method in 2D CD, which first classifies both datasets or detects the objects of interest using 3D object detection methods (Qin et al., 2015), and then compares the resulting labels of the two datasets.

***Post-refinement:*** False positives/negatives from geometric comparison occur due to artifacts of the DSM/point clouds, or incomplete 3D models. Sometimes the errors follow certain patterns, e.g. artifacts of building changes often occur at object boundaries, or in vegetation classes (due to failures of DIM methods for complex geometry). These can be well addressed if additional information/features can be extracted from the geometric or spectral data. If near-infrared data is available, NDVI (normalized difference vegetation index) can be used to eliminate disturbances from the seasonal varying vegetation. Attempt for such consideration was performed through manual interpretations (Sasagawa et al., 2008), where the radiometric difference of the images was used as a "double-check" for DSM subtraction results. To automate the process, change "candidates" can be further classified by using spectral and textural information of the original images (Fan et al., 1999; Liu et al., 2003; Pang et al., 2014).

Due to the presence of noise in DSM subtraction, some specific noise-removal approaches, for instance, morphological filtering can be used to improve the initial change masks (Chaabouni-Chouayakh et al., 2010; Choi et al., 2009; Zhu et al., 2008). External data such as map boundaries (James et al., 2012) can be used to constrain the CD masks for certain types of objects. Based on the assumption that the change maps are globally smooth, Guerin et al. (2014) applied a global optimization procedure that employs this spatial context using a generalized dynamic programming to eliminate potential inaccuracies resulting from DSM subtractions. Markov random field (MRF) approaches as used in 2D CD approaches (Diakité et al., 2014; Huang, 2013) were also developed under the 3D context (Pollard and Mundy, 2007; Qin and Gruen, 2014; Tornabene et al., 1983). Using UAV (unmanned aerial vehicle) images for CD, Qin (2014b) hierarchically refined the initial change masks using various multi-level segmentations from orthophoto and DSM. This work refined the mask using the spatial consistencies of these segments, and reported that the method can monitor even sub-building sized urban objects (such as vehicles).

The "post-refinement" approaches employ a hierarchical processing structure, where initial change evidence from geometric comparison, are refined based on available geometric and spectral features. Parameters relevant to the geometry are often easy to understand and straightforward to tune. Such "step-wise" methods are flexible to be decomposed or re-composed according to different CD applications, However, the initial CD result solely depends on the geometric comparison, and missing changes in the initial step cannot be recovered in the subsequent refinement.

***Direct feature fusion****:* Instead of processing geometric and spectral information hierarchically, the direct feature fusion simultaneously considers all channels of information. Such feature fusion can be performed in either the feature level or decision level. Although existing works in "direct feature fusion" mainly consider the fusion of multi-sources spectral data (Hartman, 2008; Longbotham et al., 2012) for CD, there are still some works that fuse both geometric and spectral information directly. Tian et al. (2013) directly fused the height and radiometric differences of Cartosat-1 datasets (only panchromatic images are available) under a change vector analysis (CVA) framework. Their subsequent work (Tian et al., 2014) adopted a Kernel Minimum Noise Fraction (KMNF) to minimize the noise statistically presented in both the height and radiometric difference for fusion, and the Iterated Canonical Discriminant Analysis (ICDA) for generating the final CD results. They reported that notable improvements were obtained in their forest CD applications using Cartosat-1 image against to methods including simple DSM differencing, CVA fusion, and other traditional classification-based methods (Remondino and El-Hakim, 2006; Wang, 2005; Young et al., 2010). Under a 3D model updating process, Qin (2014a) fused multiple change evidences resulting from DSM and spectral features via unsupervised self-organizing maps (SOM) (Kohonen, 1982; Moosavi and Qin, 2012), where the a priori information (the quality of the change evidence) can be used to weight individual change indicators for a better change determination.

Supervised approaches that use a feature vector stacking different geometric-spectral features also fall into the "direct feature fusion" category (Chehata et al., 2009; Chen et al., 2012b; Nemmour and Chibani, 2006; Pacifici et al., 2007). These methods consider both the geometric and spectral data as pure information sources. Other different kinds of information can be easily incorporated into the classifier without additional re-design of the algorithm. However, it is critical to determine the individual contribution of each source when using linear fusion models. Classifier-based models may be able to learn the weights of information sources, such as Random Forests (Breiman, 2001) and Neural Network (Foody, 1996), while this requires accurate training samples. For unsupervised fusion models (e.g. CVA), an equal contribution may not render the best results. Therefore a priori information or trial-and-error test may be needed to obtain an optimal parameter configuration.

***Classification-based method****:* Accuracy, texture and spectral differences in temporal 3D data may bring errors in direct geometric and/or spectral comparison processes. The classification-based methods propose to detect objects of interest or perform land-cover classification first, and then compare the resulting labels (classes), which avoid direct comparison of the spectral and height information. These methods share the same idea of the classification-based method as introduced in section 3.3, while the core difference is that the additional geometric information may represents a different modality data, which might have the potential to enhance both the classification and object detection accuracy for CD. A number of studies (Huang et al., 2011; Mayer, 1999; Sohn and Dowman, 2007; Zhang et al., 2015) have proven that the height information can increase the accuracy of land-cover classification to a notable level. The DSMs from LiDAR or stereo images can be essentially seen as an additional channel of information, which is equally free to be applied into popular classifiers. Researchers have investigated such a strategy via a number of classification approaches, such as in ISODATA (Olsen, 2004), maximum likelihood (Walter, 2004) decision tree (Matikainen et al., 2010), rule-based method (Champion, 2007; Olsen and Knudsen, 2005) and decision-fusion method (Nebiker et al., 2014; Rottensteiner et al., 2007).

In an urban environment, building change detection for geodatabase updating is the most relevant applications. "Building detection + Change detection" is a popular strategy to detect changes of buildings. Under this framework, Qin et al. (2015) integrated the height information to a supervised framework to detect buildings using the scanned aerial survey photos. Detected building objects from each date were then compared by considering both the height and texture dissimilarities. The height information was implemented in different levels of processing, including image segmentation, classification and CD, which was proven to be particularly effective to rebuilt buildings, as it evaluates each building object using various features such as height, texture, as well as shapes. Moreover, existing GIS data can be used as training data (Champion et al., 2009; Matikainen et al., 2010; Walter, 2004) to assist building detection and subsequently for updating. They can either be used directly as training samples (Walter, 2004), or modified using some other cues based on geometric and spectral features (Champion et al., 2009; Qin et al., 2016b). The classification method, including its 2D counterpart, is so far regarded as a popular method, since it transforms the direct geometric/spectral comparison to label changes, which offer information regarding to type changes. However, in most cases, the CD results of this method highly depend on the classification/object detection results, which it subsequently requires careful sample collection and feature design.

## 5. Summary

The creation of geodatabase (2D vector map, 3D city models) is nowadays regarded as a necessary step for initiating the digital infrastructures for smart city management. While techniques for creating engineering-grade geodatabase remain a time-consuming process, the need for updating the geodatabase in pace of the reality is at stake. This calls for efficient change detection (CD) solutions that are capable of identifying the changed area in a certain period, in order to rebuild and update the 2D/3D vector data. In the meantime, the temporal change itself acts as very valuable products for assessing dynamics of the land processes and urban development. This article provides an overview of the state-of-the-art change detection methods, structured intuitively as both 2D and 3D data change detection. Approaches of both 2D/3D CD do not necessarily separate from each other, as many methods for processing 2.5D data are indeed very similar to 2D CD methods.

CD is a disparate subject that involves many complicated issues due to the level of noises, accuracy, acquisition conditions, data format and availability of auxiliary data. Therefore, it is rather complex to conclude any approaches being superior to others. Although 3D CD being more robust than 2D CD seems to be an apparent and consistent conclusion, the existing work however still lacks good comparative study that analyzes comprehensively the performance of spectral and geometric data for CD under different conditions and scenarios, including comparison using similar data under similar application contexts. However, so far we have a general understanding on the prevalent methods and their performances, including pixel-based, object-based and classification-based methods. It seems an ad-hoc CD method for applications are still the mainly trend, particularly for the task of updating geodatabases, where the availability of source data (images and LiDAR point clouds) and the format of vector data may vary greatly. In the case that semantic information of the geodatabase is available, it is recommended to incorporate information such as building block number, streets, number of floors, types of objects, and types of buildings into the CD algorithm to render more robust/engineered solutions.

As we look into the CD algorithms themselves, "**comparison** of data for changes on **object of interest**" are the keys. These seemingly independent keywords are in a great connection to three fundamental issues:

1) advanced spectrum correction and comparison methods;

2) advanced 3D data generation algorithms that provide more accurate 3D measurements for accurate geometric comparison;

3) high-level feature extraction and machine-learning methods for detecting the object of interest.

Therefore, in addition to the efforts in reducing noises and unwanted changes, the advanced CD algorithms rely on the future endeavor of these three aspects to put forth more integrated and robust methods for geodatabase updating.

This is a pre-publication version of the published book chapter in 3D/4D City Modelling: from sensors to applications. The final contents are subject to minor edits

This is a pre-publication version of the published book chapter in 3D/4D City Modelling: from sensors to applications. The final contents are subject to minor edits